\documentclass[10pt,twocolumn,letterpaper]{article}

\usepackage{cvpr}              
\usepackage{booktabs}
\usepackage{multirow}
\usepackage{tabularx} 
\usepackage{afterpage}
\usepackage{float}
\usepackage[accsupp]{axessibility} 
\usepackage[dvipsnames]{xcolor}

\usepackage{graphicx}
\usepackage{ulem}

\definecolor{cvprblue}{rgb}{0.21,0.49,0.74}
\usepackage[pagebackref,breaklinks,colorlinks,citecolor=cvprblue]{hyperref}

\def\paperID{13364} 
\def\confName{CVPR}
\def\confYear{2024}
\title{WaveMo: Learning Wavefront Modulations to See Through Scattering}

\author{
    Mingyang Xie$^{1}$\thanks{Equal Contribution.} \quad Haiyun Guo$^{2}$$^{*}$ \quad Brandon Y. Feng$^{3}$ \quad Lingbo Jin$^{2}$ \\ 
    Ashok Veeraraghavan$^{2}$  \quad Christopher A. Metzler$^{1}$\\\\
    $^{1}$University of Maryland \qquad $^{2}$Rice University  \qquad $^{3}$Massachusetts Institute of Technology
}

\begin{document}
\maketitle

\begin{abstract}
Imaging through scattering media is a fundamental and pervasive challenge in fields ranging from medical diagnostics to astronomy.
A promising strategy to overcome this challenge is wavefront modulation, which induces measurement diversity during image acquisition. 
Despite its importance, designing optimal wavefront modulations to image through scattering remains under-explored.
This paper introduces a novel learning-based framework to address the gap.
Our approach jointly optimizes wavefront modulations and a computationally lightweight feedforward ``proxy'' reconstruction network.
This network is trained to recover scenes obscured by scattering, using measurements that are modified by these modulations.
The learned modulations produced by our framework generalize effectively to unseen scattering scenarios and exhibit remarkable versatility.
During deployment, the learned modulations can be decoupled from the proxy network to augment other more computationally expensive restoration algorithms. 
Through extensive experiments, we demonstrate our approach significantly advances the state of the art in imaging through scattering media. 
Our project webpage is at \href{https://wavemo-2024.github.io/}{https://wavemo-2024.github.io/}.
\end{abstract}

\section{Introduction}
\label{sec:intro}
Imaging through scattering media presents a significant challenge across diverse scenarios, ranging from navigating with fog~\cite{Ren2016SingleID, Ancuti2020NHHAZEAI, shi2022seeing}, rain~\cite{zhang2023learning, yu2022towards, Fu2017RemovingRF}, or murky water to recovering intricate structures through human skin and tissue~\cite{gigan2022roadmap,yoon2020deep,watnik2018wavefront}. 
The core of this challenge are the irregular phase delays light experiences as it scatters. These phase delays blur and warp any images captured through
the scattering media. Effectively addressing this issue is crucial for unlocking new computer vision capabilities in fields such as medical imaging and astronomy.

\begin{figure}[t]
  \centering
   \includegraphics[width=\linewidth]{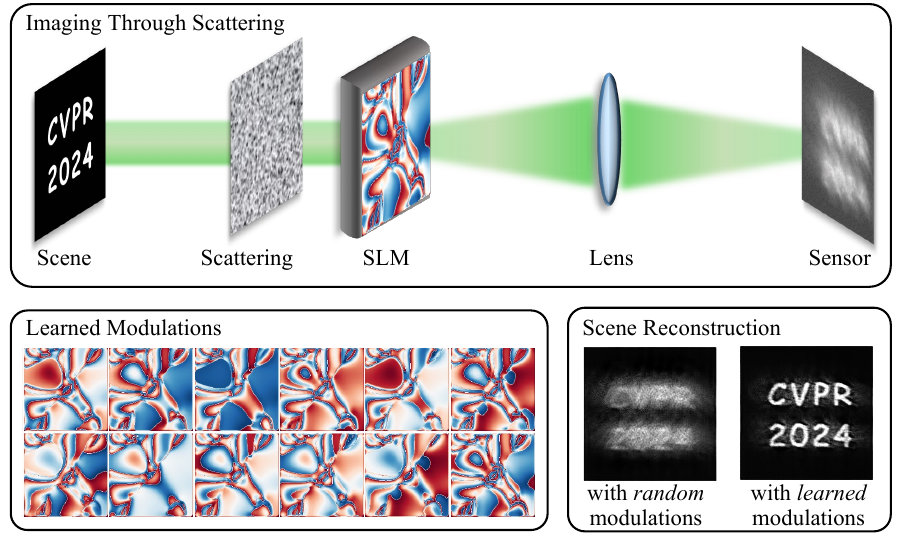}
   \caption{\textbf{Learned Wavefront Modulations}. {\it Top}: During acquisition, we can modulate the wavefront of scattered light by using a spatial light modulator (SLM), and capture a set of image measurements useful for scene reconstruction.
   {\it Bottom left}: We propose learning modulations that enhance our ability to recover the scattered scene.
   {\it Bottom right}: Our learned modulations drastically improve the reconstruction quality of a state-of-the-art method~\cite{feng2023neuws} that previously applies randomly chosen modulations.}
   \label{fig:onecol}
   \vspace{-5pt}
\end{figure}

\begin{figure*}[t]
    \centering
    \includegraphics[width=0.97\textwidth]{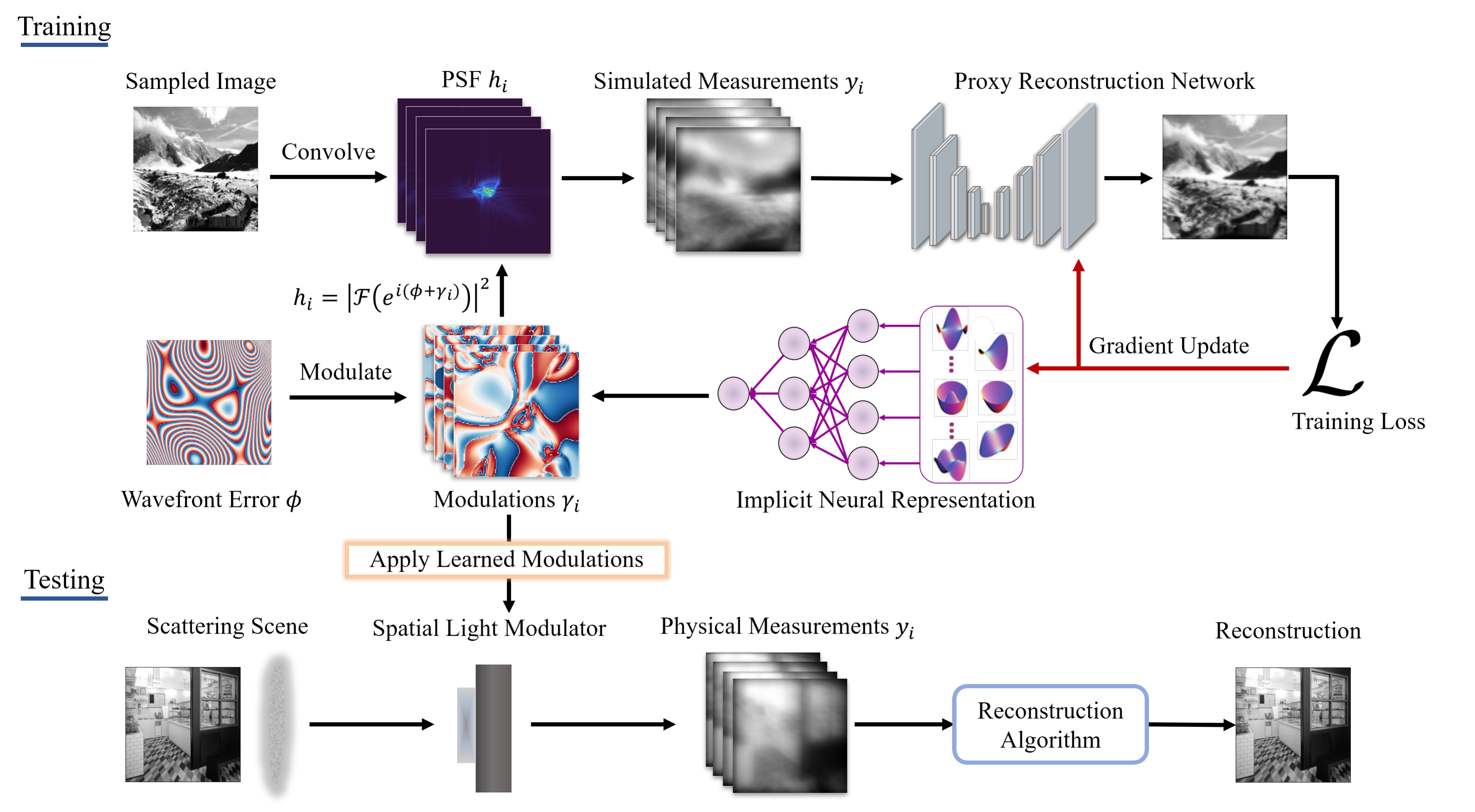} 
    \caption{\textbf{Overview of the Proposed End-to-end Learning Framework.} During training, we jointly optimize the implicit neural representation for the wavefront modulations and the proxy reconstruction network in an end-to-end fashion. During inference, one can apply the learned wavefront modulations to any reconstruction algorithms for imaging through scattering, either a trained feed-forward reconstruction network or an unsupervised iterative optimization algorithm~\cite{feng2023neuws}. The former approach is far faster and performs better when test data and training data fall into similar distributions, while the latter generalizes better to unseen distributions of target scenes.}
    \label{fig: overview}
\end{figure*}

From a frequency domain perspective,  the effect of a scene passing through scattering is the destruction or filtering of frequency content. 
Recovering these lost frequencies is very difficult using purely computational methods, which are effectively forced to speculate on the missing frequencies based on those that have been preserved. 
A question naturally arises: Can we modify the measurement process to prevent the loss of information during acquisition? 

A promising solution is to actively modulate the scattered wavefront during acquisition, in hopes that applying multiple modulations increases the chance that a frequency component is preserved.
This approach is related to phase diversity imaging~\cite{gonsalves1982phase,paxman1992joint,vogel1998fast,thelen1999maximum,kner2013phase,xin2019object,kang2023coordinate,reiser2023phase}, which typically captures multiple scattered images of a static scene with different modulations applied using deformable mirrors or spatial light modulators (SLM).
However, phase diversity imaging traditionally was limited to imaging simple scenes with mild optical aberrations, and the phase modulations were often selected based on simple heuristics.
The recent development of neural wavefront shaping (NeuWS)~\cite{feng2023neuws} brings a significant breakthrough by leveraging recent innovations in machine learning and enables imaging high-resolution and dynamic scenes through severe aberrations.
Still, the wavefront modulations applied by NeuWS during acquisition are chosen randomly without explicit consideration of their impact on frequency preservation.

In this paper, we take a significant step towards a more systematic and principled approach in designing effective modulations that better preserve frequencies against scattering during image acquisition. 
Moving beyond the traditional reliance on simple heuristics for determining modulation patterns, our method combines the strengths of differentiable optimization and data-driven learning.
Central to our approach is the novel integration of the optical model of wavefront modulations with a {\it proxy} reconstruction network, resulting in a fully differentiable system.
This integration allows for the simultaneous optimization of both the proxy reconstruction network and the modulation patterns, which are optimized end-to-end based on a large, simulated training dataset.
While imaging through scattering often requires iterative optimization algorithms because data-driven feedforward networks often fail to generalize outside their training domain, this paper demonstrates the remarkable finding that training a {\it hard-to-generalize} network can become a useful proxy allowing us to learn {\it generalizable} modulations.
Crucially, this proxy feedforward network allows us to bypass the expensive backpropagation computation through an iterative reconstruction algorithm, which would be necessary, but hopelessly impractical, if we were to naively optimize the modulations: each iteration would require us to finish an entire iterative reconstruction algorithm.
As will be shown in our simulated and real data experiments, our learned modulations significantly enhance the reconstruction capability of both our data-driven proxy network and a state-of-the-art iterative optimization-based method~\cite{feng2023neuws} for imaging through scattering.

Our contributions are:
\begin{itemize}[leftmargin=.25in]
  \item We present a novel end-to-end learning framework to optimize acquisition-time wavefront modulations to enhance our abilities to see through scattering.
  \item We demonstrate in both simulated and real experiments that our learned modulations substantially improve image reconstruction quality and effectively generalize to unseen targets and scattering media.
  \item We show that the learned modulations can be decoupled from the jointly-trained proxy reconstruction network and significantly enhance the reconstruction quality of state-of-the-art unsupervised approaches.
  \end{itemize}

\section{Related Work}
\label{sec:related_work}

\paragraph{Learning-based Imaging Through Scattering Media.}
A detailed review of imaging through scattering media can be found in~\cite{gigan2022roadmap}. We highlight a few recent learning-based approaches. \textcolor{black}{Various data-driven approaches can be applied to image through scattering~\cite{li2018deep,DeepInverseCorr,tahir2022adaptive,hu2023adaptive,chen2023imaging,sun2018efficient}}. 
They generally capture or synthesize large quantities of training data and then learn a mapping from corrupted measurements to reconstructed scenes. 
However, they tend to struggle with out-of-distribution data. 
There have also been unsupervised algorithms by fitting network-based parameterizations of the scene to a collection of measurements~\cite{feng2023neuws,kang2023coordinate}, which can generalize across different distributions of aberrations and target scenes. Both of these methods take advantage of phase diversity.

\paragraph{End-to-End Learning of Optical Systems.}
End-to-end learning is a flexible learning-based system design framework that describes optical systems using parameterized differentiable optical models which can then be optimized with training data~\cite{Sitzmann:2018}. 
End-to-end learning has been applied successfully for extended depth-of-field imaging~\cite{Sitzmann:2018, Ikoma:2021, Liu:22,shah2023tidy}, depth estimation~\cite{phasecam, Chang:2019:DeepOptics3D, Ikoma:2021,shah2023tidy}, high-dynamic-range imaging~\cite{metzler2020deep,sun2020learning}, \textcolor{black}{seeing through near-lens occlusions~\cite{shi2022seeing}}, and many other applications. To facilitate end-to-end learning, prior works have also incorporated the usage of a proxy network to mimic the response of a specific optical system or black-box algorithm~\cite{Tseng2019HyperProxy, peng2020neural}. 

\paragraph{Phase Diversity Imaging.}
Phase diversity imaging (PDI) can recover a clear image of an aberrated scene by obtaining images with deliberately introduced known aberrations. 
PDI does not pose strong assumptions on the target \cite{xin2019object} or the aberration \cite{kendrick1994phase}, and the optics and calibration process required are relatively simple compared to other adaptive optics imaging \cite{gonsalves1982phase}. 
While early applications of PDI are often constrained to simple aberrations due to computation complexities \cite{smith2012fast}, the introduction of coded diffraction patterns has enhanced its capacity for reconstructing intricate scenes \cite{candes2015phase}. 
Recent PDI developments include improved wavefront sensing~\cite{zhang2017high, xin2019object}, phase retrieval~\cite{reiser2023phase, xiang2022phase}, adaptive optics~\cite{echeverri2016vortex, kang2023coordinate}, and imaging through scattering~\cite{feng2023neuws, chen2022enhancing, yeminy2021guidestar}.

\section{Learned Wavefront Modulations}
\label{sec:method}
Here we describe our approach for learning wavefront modulations through training a proxy reconstruction network.



\subsection{Scattering Problem}
This paper focuses on the scenario where capturing an image through scattering media can be modeled as
\begin{align}\label{eqn:forwardmodel}
	y = h(\phi)*x+\epsilon,
\end{align}
where $y$ is the captured measurement, $x$ is the target scene that we aim to reconstruct, $\epsilon$ is noise, and  $h(\phi)$ is the unknown, spatially invariant point spread function (PSF) describing the optical scattering, which manifests as unknown phase delays $\phi$ to the wavefront.
Assuming the target object is illuminated with spatially incoherent monochromatic light, the PSF $h$ is related to the phase error $\phi$ by 
	$h(\phi)=|\mathcal{F}(m\circ e^{j(\phi)})|^2$,
where $\mathcal{F}$ is the 2D Fourier transform, $\circ$ is the hadamard product and $m$ is the aperture mask~\cite{GoodmanFourierOptics}. Our goal is to recover $x$ from $y$ captured through $\phi$. 

\subsection{Phase Diversity}
Phase diversity imaging seeks additional information about the phase error $\phi$ by capturing a sequence of measurements $y_1$, $y_2$, ... $y_K$, each with additional {\em known} phase errors $\gamma_1$, $\gamma_2$, ..., $\gamma_K$. 
As a result of these additional modulations, the imaging forward model becomes $y_i =h(\phi+\gamma_i)*x + \epsilon_i$.
In traditional approaches, the phase modulations $\{\gamma_i\}_{i=1}^{n}$ were typically either simple defocus sweeps or randomly chosen patterns generated from Zernike polynomials~\cite{paxman1992joint,feng2023neuws},
and their reconstruction objective seeks to minimize 
\begin{align}\label{eqn:loglike_2}
\mathcal{L}_{PDI}(\hat{x},\phi)=\sum_{i=1}^K {\big \|}y_i-h(\phi+\gamma_i)*\hat{x}{\big \|}^{2},
\end{align}
over $\hat{x}$ and $\phi$, where $\hat{x}$ is the estimate of scene $x$, assuming $\epsilon_i$ follows an i.i.d.~additive Gaussian distribution. 

\subsection{Learned Modulations}
While previous works have leveraged differentiable proxies of physical systems to design better algorithms~\cite{Tseng2019HyperProxy,peng2020neural}, this work leverages a differentiable proxy, $\mathcal{P}$, of non-differentiable algorithms to design better physical systems---specifically the phase modulation $\Gamma = \{\gamma_{i}\}_{i=1}^{K}$ used to image through scattering.

We sample a training image $x_n$ and simulate passing it through an optical aberration drawn from some known distribution while we apply $K$ learnable modulations to the scattered wavefront. This model allows us to simulate a collection of $K$ images $Y_n^{\Gamma} = \{y_1^{n},y_2^{n},...y_K^{n}\}$, which are different scattered observations of $x_n$ based on Eq.~\eqref{eqn:forwardmodel}.

We then optimize both the proxy reconstruction algorithm $\mathcal{P}$ and the modulation patterns $\Gamma$ to minimize
\begin{align}\label{eqn:MSEboth}
\mathcal{L}_{learned}(\mathcal{P}, \Gamma) = \sum_{n=1}^{N} \|\mathcal{P}(Y_n^{\Gamma}) - x_{n} \|^{2}.
\end{align}



Note that the ultimate goal of the learning is not to design a set of modulations specific to the network $\mathcal{P}$. Rather, we use the performance of $\mathcal{P}$ as a proxy to probe how effective the modulations $\Gamma$ are for the scattering problem. The differentiability of $\mathcal{P}$ allows for back-propagation to guide the optimization of the modulations.
As we will demonstrate, while $\mathcal{P}$ itself may not generalize outside of its training data domain, the learned modulations $\Gamma$ are compatible with other more generalizable reconstruction algorithms.

Our end-to-end training pipeline is illustrated in Figure~\ref{fig: overview}.
For the proxy reconstruction network, we choose a U-Net with skip connections and self-attention modules~\cite{Oktay2018AttentionUL}.
To regularize the highly non-convex problem of optimizing modulations, we use an implicit neural representation for the modulations $\Gamma$. 
Specifically, the MLP ${G}$ takes a fixed vector $Z$ with 28 channels, which corresponds to the first 28 Zernike polynomials \cite{lakshminarayanan2011zernike}; it outputs a 16-channel vector corresponding to 16 modulation patterns: $\Gamma_{G}=G(Z)$.

Incorporating the implicit neural representation for the modulations, our final loss function becomes
\begin{align}
\label{eqn:loglike_1}
\mathcal{L}_{final}(\mathcal{P}, G)= \sum_{n=1}^N {\big \|} \mathcal{P}(Y_n^{\Gamma_{G}}) -x_{n}{\big \|}^{2},
\end{align}
where $Y_n^{\Gamma_{G}} = \{h(\phi_n + G(Z)_i) * x_n\}_{i=1}^{K}$.

\subsection{Applying Learned Modulations}
While the trained proxy network $\mathcal{P}$ demonstrates considerable capability in recovering scenes from scattering in our experiments, supervised data-driven methods often fail to generalize to novel or cross-domain scenes, particularly when compared to unsupervised approaches.

Therefore, after obtaining the learned modulations ${\Gamma_{G}}$ and applying them during real-world acquisition, we send the resulting modulated measurements to an unsupervised iterative optimization-based reconstruction algorithm~\cite{feng2023neuws} which does not suffer from the generalization issue of data-driven methods. 
In effect, our method offers a best-of-both-worlds scenario:
the modulations ${\Gamma_{G}}$ are effectively learned thanks to the joint supervised training with the proxy network $\mathcal{P}$, but their enhanced data acquisition quality is transferrable to other reconstruction algorithms, thus allowing us to benefit from the generalization and domain adaptability inherent in unsupervised reconstruction methods.
\section{Analysis of Learned Modulations}
\label{sec:analysis_on_simulation}
Before we show the performance of the learned modulations with real-world experiments in Section~\ref{sec:experiments}, 
we first analyze the effectiveness of learned modulations in simulation. 


\subsection{Impact on Frequency} \label{sec:sim_freq}

\begin{figure}[t]
    \centering
    \includegraphics[width=0.4\textwidth]{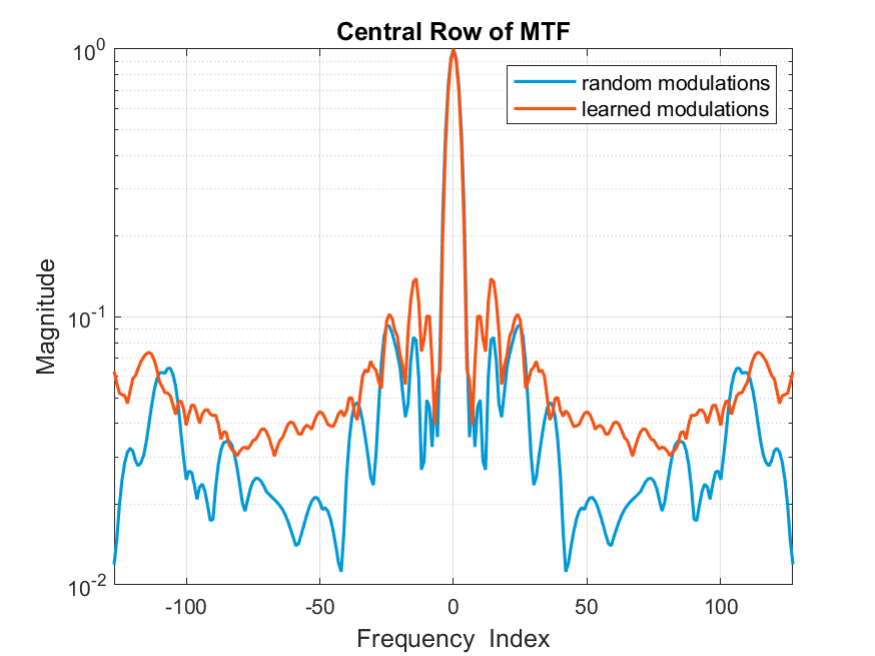} 
    \caption{\textbf{MTF Comparison}. We compare the MTF of learned wavefront modulations vs. the MTF of modulations randomly drawn from the same distribution of the scattering media. The X-axis represents spatial frequency, and the Y-axis represents the modulation transfer (the higher the better). Our learned wavefront modulations exhibit a higher MTF compared to unoptimized ones, especially over higher frequency bands, suggesting the learned modulations preserve more high-frequency information. }
    \label{fig: mtf}
\end{figure}
When the scattering media scrambles the incoming wavefront, it destroys some of the target scene's frequencies, either by destructively interfering with them or by scattering them outside the imaging sensor. If those frequencies are not captured in the measurement, it becomes difficult to faithfully recover them.
By applying multiple learned wavefront modulation patterns, we increase the chance that each frequency is better preserved by at least one of the modulations.
With the extra information from all modulated measurements, a better image reconstruction performance could be achieved regardless of the reconstruction algorithm.

\begin{figure}[t]
    \centering
    \includegraphics[width=0.42\textwidth]{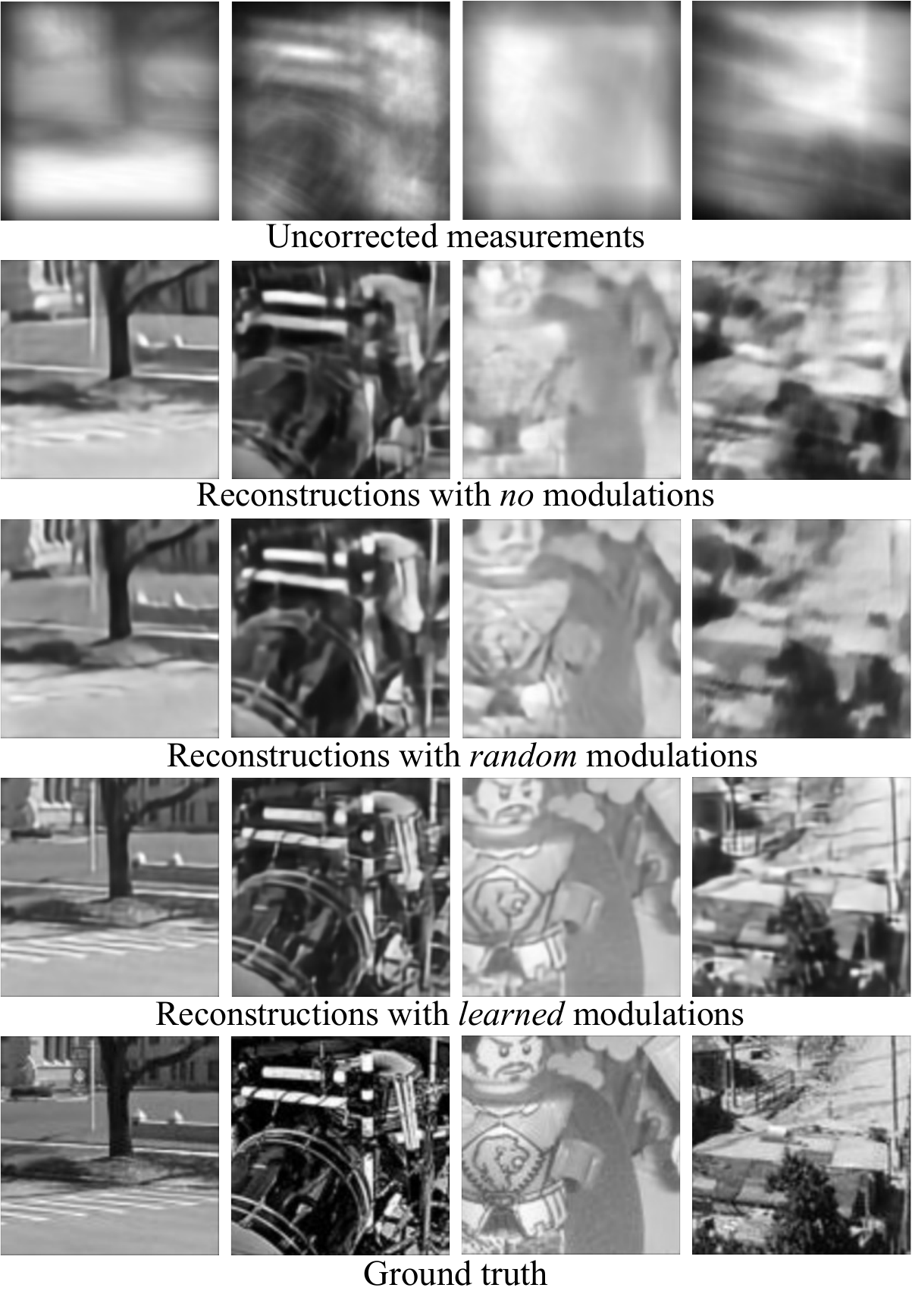} 
    \vspace{-5pt}
    \caption{\textbf{Proxy Network Reconstruction on Simulated Scattering}. Simulated imaging results through different aberrations. Learned modulations lead to a better quality.}
    \label{fig: sim}
\end{figure}

Thus, we assess the performance of our learned modulations by their combined ability to preserve frequencies.
We choose to the commonly used Modulation Transfer Function (MTF) as our metric~\cite{mtf_1999, mtf_2022vivo, mtf_holography, mtf_tian2015, mtf_wolgang2021}.
MTF measures the contrast produced by an imaging system at different spatial frequencies, defined as $\text{MTF}_{i}^{\omega}(\phi) = {\big |} \mathcal{F}[h_i(\phi)] {\big |}^{\omega}$,
where $h_i$ is the PSF for the $i$-th modulation and the superscript $\omega$ denotes indexing the MTF at a specific frequency $\omega$. For the multiple modulations in our problem setup, we define the combined MTF as:
\begin{align}\label{eqn:mtf}
\text{MTF}_{\text{comb}}^{\omega}(\{\gamma_i\},\phi) = \max_{i} \text{MTF}_{i}^{\omega}(\phi+\gamma_i),
\end{align}
where for each frequency $\omega$, we take the maximum over the MTFs calculated from all modulations.

In Figure~\ref{fig: mtf}, we plot the combined MTF for both the learned wavefront modulations and the random ones as baseline. 
The learned ones have a clear advantage at high frequencies (the higher the frequency is, the further away it is to the center of the X-axis), suggesting that the learned modulations are better at preserving finer details of the target scene. Note that during training, we never explicitly encourage the MTF to be higher ---  the only loss term we use is the mean square error between reconstruction and ground truth image. This implies that our end-to-end training implicitly encourages the modulations to preserve higher-frequency information.

\subsection{Impact on Reconstruction} \label{sec:sim_recon}
We evaluate our jointly learned modulations and proxy network both quantitatively and qualitatively. Both the training and test set are from the {MIT Places 365 Dataset}~\cite{Zhou2018PlacesA1}. Training details are provided later in Section~\ref{sec:hardware}. Here, we use two baselines: (1) simply training a feed-forward reconstruction network without any wavefront modulations, and (2) training a feed-forward reconstruction network with randomly chosen, fixed wavefront modulations as used in prior PDI work. For a fair comparison, our approach and baselines use the same network architecture and are trained on the same data for the same number of iterations. Their only difference is the usage of wavefront modulations.

As shown in Figure~\ref{fig: sim}, compared to baselines, reconstructions using learned modulations demonstrate finer details and higher consistency with the ground truth. Table~\ref{tab:sim_metric} shows the quantitative comparison averaged over a test set of 1,000 images, where our approach outperforms the baselines by over 3.9 dB.
While the results above are all from the jointly trained proxy reconstruction network, we will show in Section~\ref{sec:experiments} that our learned wavefront modulations can decouple from the proxy network and also improve reconstruction quality based on iterative optimization~\cite{feng2023neuws}.

\begin{figure*}[t]
  \centering
   \includegraphics[width=0.95\linewidth]{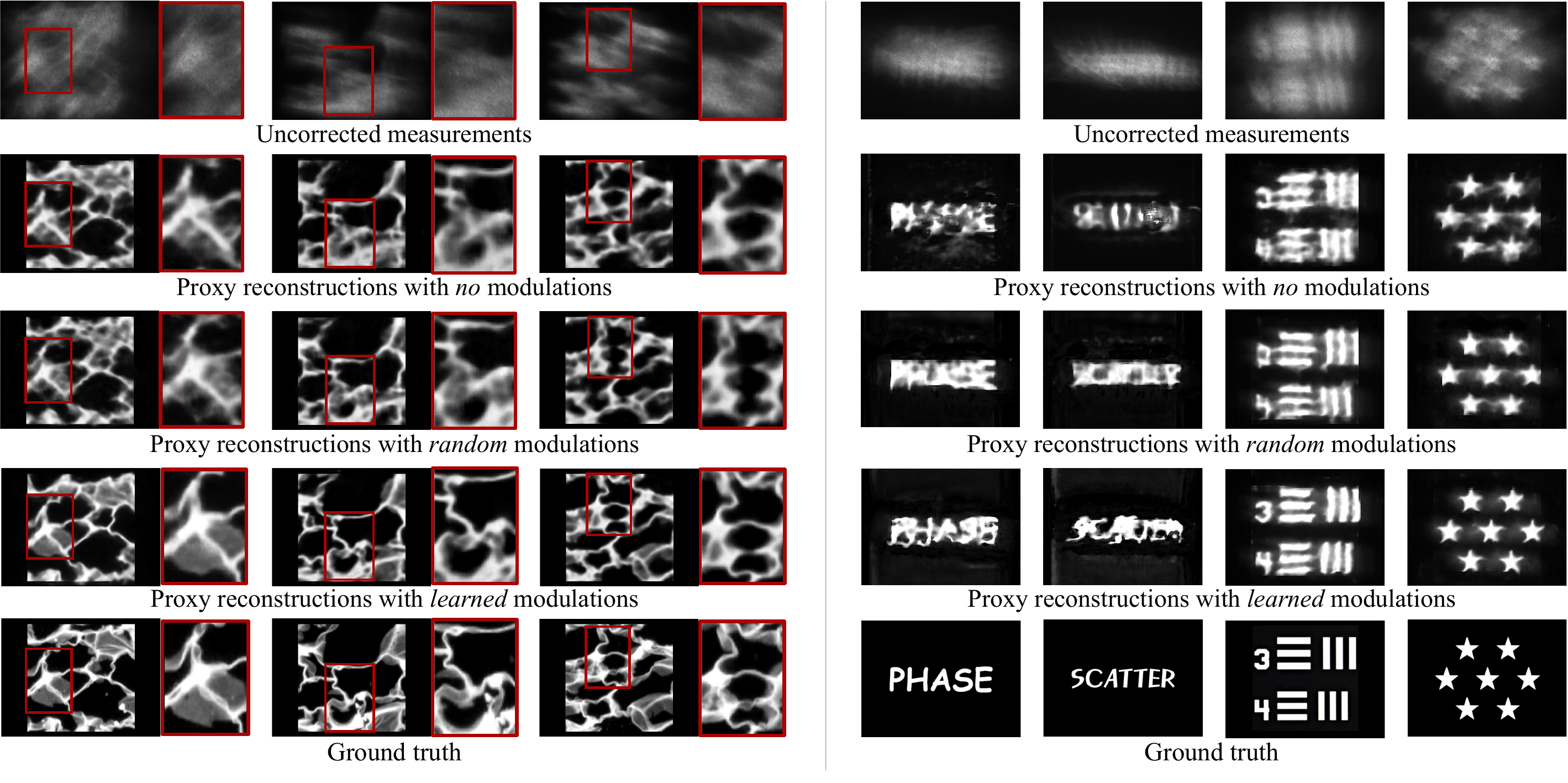} 
   \caption{\textbf{Proxy Network Reconstruction on Physical Scattering}. Experimental results of imaging objects through scattering media by using our proxy reconstruction network. The left columns are in-distribution adipose tissue slides (zoomed-in region labeled with red boxes); the right columns are out-of-distribution targets. Learned modulations yield superior imaging quality for both in-distribution and out-of-distribution scenes, with the former out-performing the latter.}
    \label{fig: feedfoward}
\end{figure*}

\begin{table}
\captionsetup{skip=5pt}
    \vspace{-5pt}
    \centering
    \begin{tabular}{lcccc}
    \toprule
         \multirow{2}{*}{Metric} & \multicolumn{3}{c}{Modulations} \\
         \cline{2-4}\noalign{\vspace{1pt}} & None & Random & Learned \\
    \midrule
         PSNR & 25.476  & 26.439 & \textbf{30.391} \\
         SSIM & 0.7640 & 0.7980 & \textbf{0.9082} \\
    \bottomrule
    \end{tabular}

    \caption{\textbf{PSNR and SSIM of Simulation Results.} We report the test-time performance of our jointly trained modulations and the proxy network. We compare it against two baselines where either we do not use wavefront modulations or use randomly chosen, unoptimized modulations. The metrics are averaged over the reconstruction of 1000 test images, each of which is scattered with a randomly sampled optic aberration that is unseen during training. The results demonstrate the huge performance improvement from the optimization of wavefront modulations.}
    \label{tab:sim_metric}
    \vspace{-5pt}
\end{table}

\section{Implementation}
\label{sec:hardware}
\paragraph{Software Implementation \& Training.} 

For the proxy reconstruction network, we use an attention U-Net~\cite{Oktay2018AttentionUL}; for the MLP that represents the 16 wavefront modulations, we use a two-layer MLP with leaky-ReLU activation. In each training iteration, the optical aberrations are sampled randomly from Zernike basis functions. The standard deviation of the coefficients for each Zernike basis is randomly generated from a uniform distribution from 5 to 6. We train our proposed framework for 2 million iterations on the {MIT Places Dataset}~\cite{Zhou2018PlacesA1} with the Adam Optimizer and a learning rate of 0.001. Training took 12 hours on an NVIDIA RTX A6000 GPU. For experimental results, we finetune our trained model on the captured bio-tissue data under the same training settings except with a learning rate of 0.0001. 

\begin{figure}[t]
    \centering
    \includegraphics[width=0.47\textwidth]{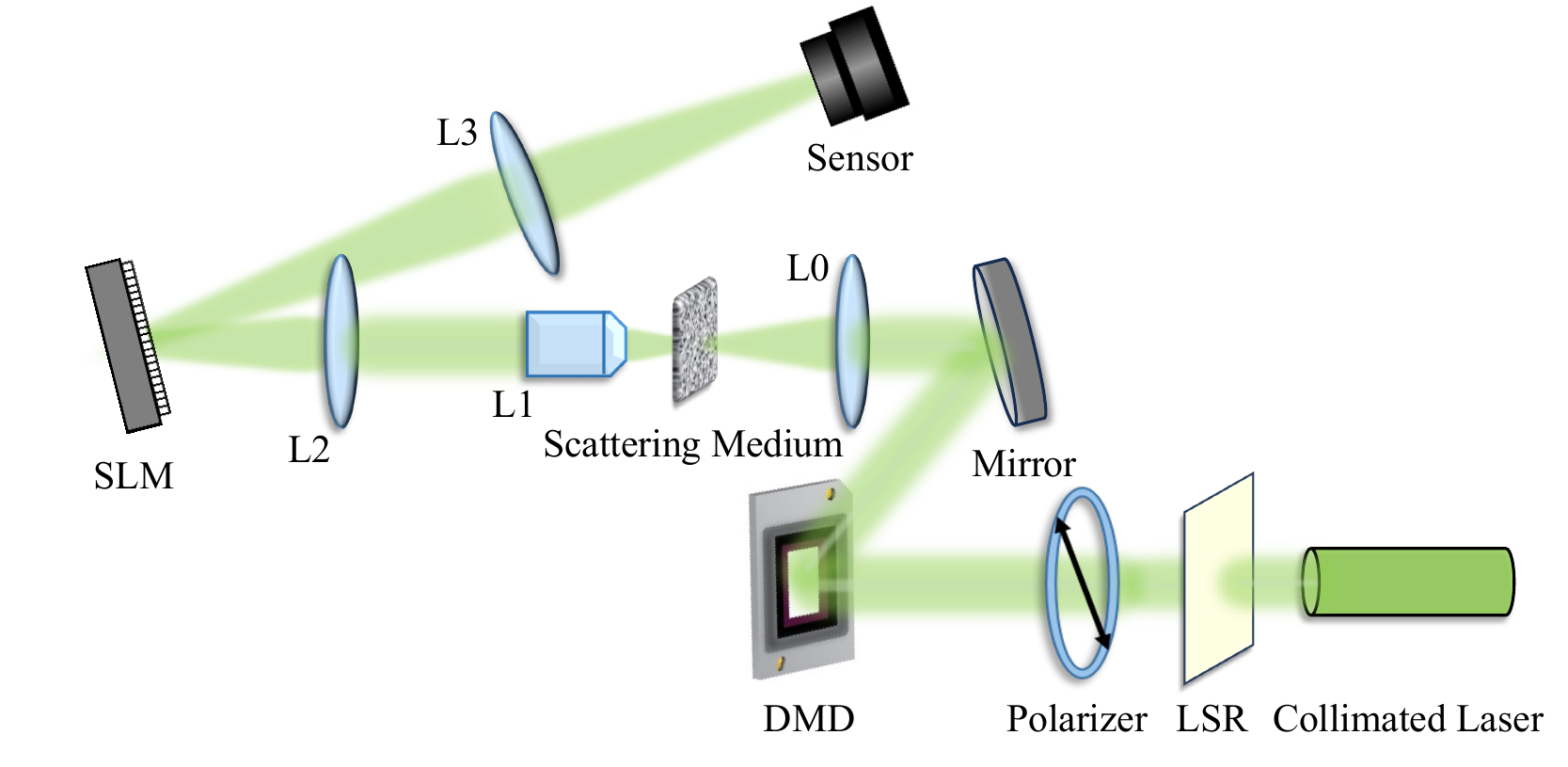}
    \caption{\textbf{Optical Setup.} A spatially incoherent light source illuminates the DMD and passes through thick scattering. The DMD displays the target images for training and testing. The SLM is placed onto the Fourier plane and projects learned patterns. The wavefront is modulated and subsequently imaged onto a camera.}
    \label{fig: optical setup}
\end{figure}

\vspace{-10pt} 
\paragraph{Hardware Implementation.}  Our optical configuration is depicted in Figure \ref{fig: optical setup}. A continuous-wave laser with a wavelength of 532 nm was collimated and subsequently passed through a laser speckle reducer (Optotune LSR-4C-L). The resulting spatially incoherent light was then polarized and reflected by a digital micromirror device (DMD). The DMD (LightCrafter™ 4500) display is capable of rapidly presenting 8-bit target images. For ease of assembly, a reflecting mirror was incorporated after the DMD. Then, the light is focused by lens L0 ($f=100$mm) onto the front focal plane of L1, a 10$\times$ microscopic objective.  \textcolor{black}{Our optical aberration---resin painted on glass---was placed at the same location.} L1 and a tube lens L2 ($f=200$mm) constitute a 4$f$ system, and thereby the SLM is manipulating the Fourier plane. We used a HOLOEYE LETO-3 phase-only SLM to modulate the wavefront. Finally, the light reflected by the SLM was imaged on the camera (1384$\times$1036 pixel Grasshopper3) via L3 ($f=500$mm). 
Our capture speed is limited to 30 fps by the camera frame rate. 
\begin{figure*}[t]
    \centering
    \includegraphics[width=0.95\textwidth]{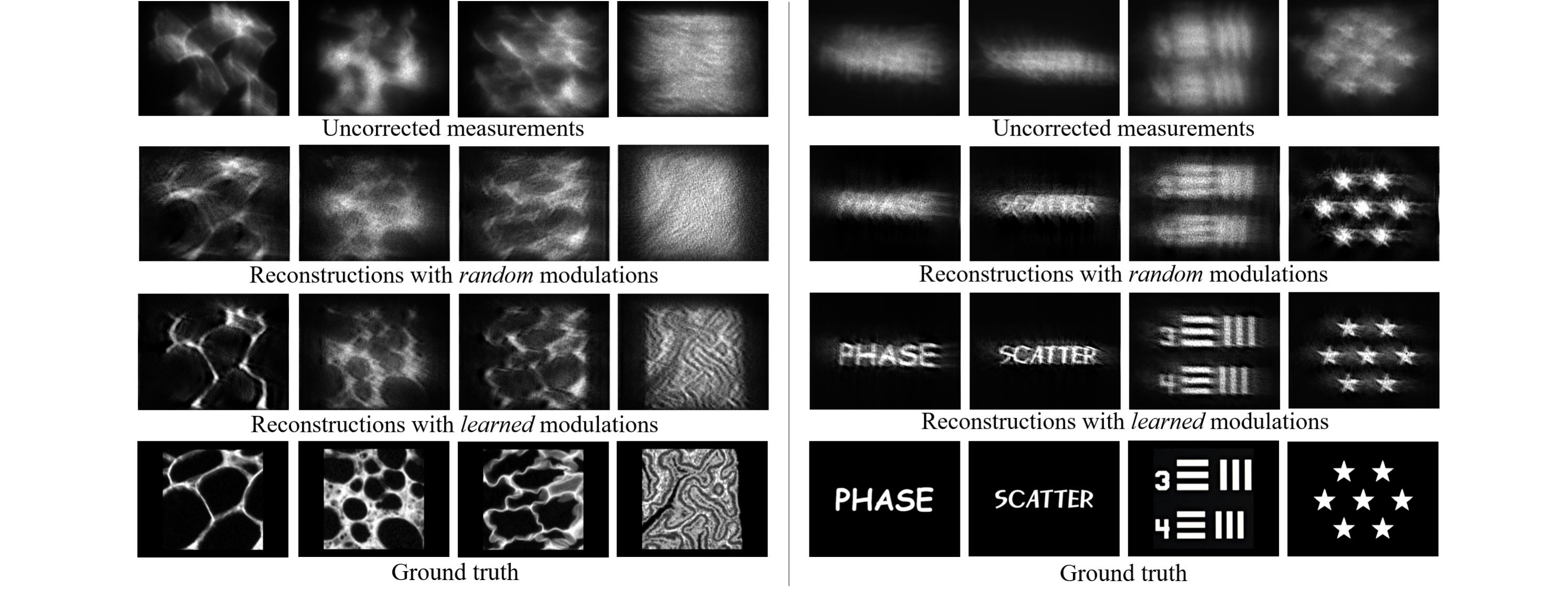} 
    \caption{\textbf{Unsupervised Reconstruction on Physical Scattering with Static Scenes}. Experimental results of imaging different static targets through scattering media using learned wavefront modulations with unsupervised iterative optimization~\cite{feng2023neuws}. Learned modulations enhance reconstruction quality.}
    \label{fig: NeuWS_static}
\end{figure*}
\vspace{-10pt} 
\paragraph{Experimental Dataset.}  Our experimental data are from human pathological slides, imaged with optical microscopy (Zeiss Microscope Axio Imager.A2) and a 10$\times$ objective. Each mage is 256$\times$256 pixels. Our dataset includes 1,000 sectional images of adipose tissues. 40 of these images were utilized for testing and the rest were used for training. To evaluate the generalizability across different human tissues, we also imaged sections of human stomach and glandular epithelium, trachea, and finger. Additionally, we tested with targets outside the domain of the training data, such as numbers and letters. 



\begin{figure*}[ht]
    \centering
    \includegraphics[width=0.95\textwidth]{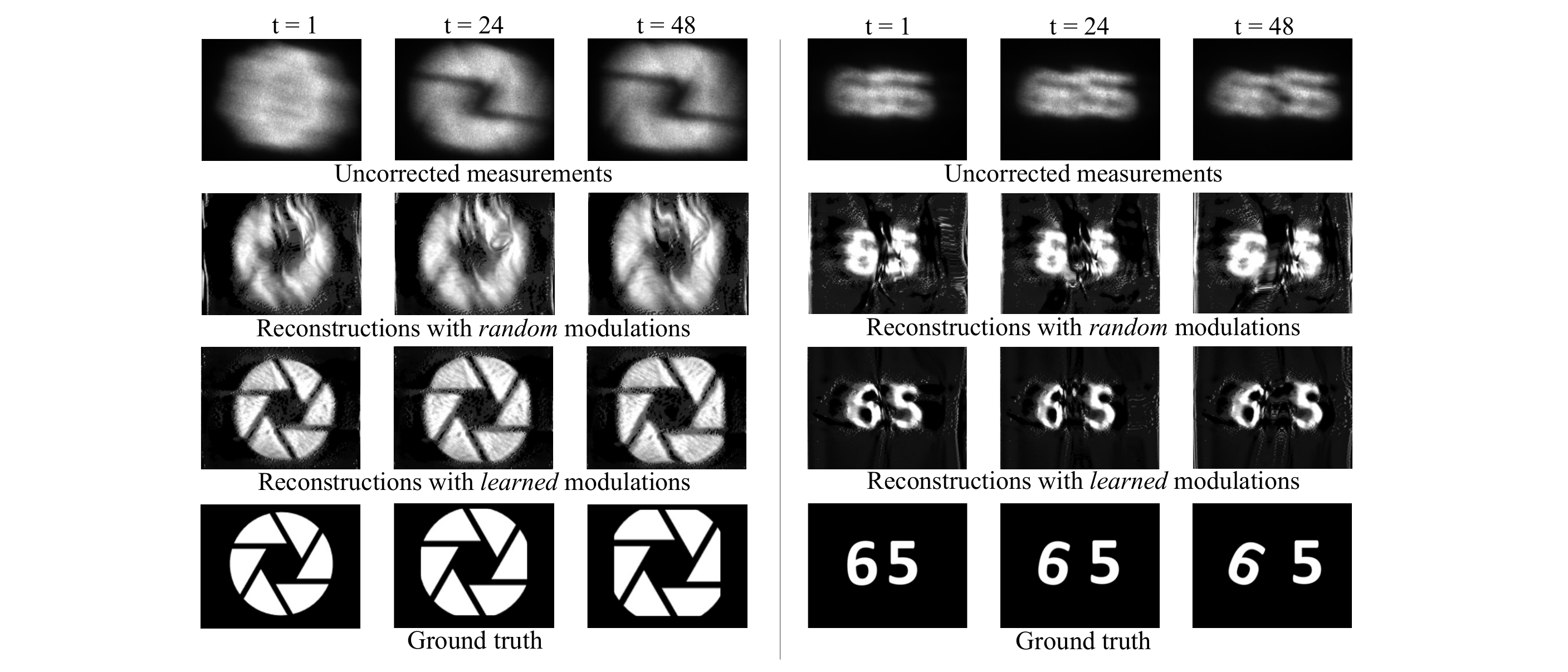} 
    \caption{\textbf{Unsupervised Reconstruction on Physical Scattering with Dynamic Scenes.} Experimental results of imaging an opening camera aperture through severe aberrations (left) and imaging non-regional motions (number 6 rotating clockwise and number 5 translating to the right) through severe aberrations. We use the unsupervised iterative reconstruction approach~\cite{feng2023neuws}. Learned modulations highly improve the quality of the reconstructed dynamic scene.}
    \label{fig: NeuWS_dynamic}
\end{figure*}

\section{Experimental Results}
\label{sec:experiments}

We employed the optical system depicted in Figure \ref{fig: optical setup}. Our learned modulations significantly enhance the imaging capabilities through scattering media. 
We tested on datasets from both similar and distinctly different distributions from our experimental training data. Our learned modulations consistently yield superior visual reconstruction results. 
Furthermore, we evaluated these modulations on both static and dynamic scenes. 
We show that our learned modulations significantly enhance reconstruction in both cases, which suggests a broad applicability of our approach.


\subsection{Proxy Reconstruction Network}

The proxy network, fine-tuned on 960 adipose tissue images through random scattering, was evaluated on a test set from the same tissue type, as shown in the left of Figure \ref{fig: feedfoward}. Reconstructions driven by measurements of 16 learned SLM patterns preserved the original contrast and high-frequency details of the dyed sections. The close-ups show remarkable detail accuracy. By comparison, reconstructions trained on measurements of random or no modulation struggled, particularly in lower-intensity areas, yielding blurry outputs. 
Beyond achieving good reconstruction fidelity on objects similar to our training dataset, we extended our evaluation to vastly different data distributions, like numbers and letters. The results, as shown on the right of Figure \ref{fig: feedfoward}, clearly indicate that the learned modulations yield better reconstruction quality, particularly in the USAF target and stars. PSNR and SSIM are computed in Tables \ref{tab:exp_psnr} and \ref{tab:exp_ssim}. 
Among the table cells, ``None'' means that the measurements for each target only comprised the unmodulated image (no modulation patterns are applied); ``Random'' means that the modulation patterns are randomly sampled from the same distribution as the aberrations. 

\subsection{Unsupervised Iterative Approaches}
While the proxy network $\mathcal{P}$ exhibits considerable effectiveness in real-world experiments, its performance remains unsatisfactory for out-of-distribution data, which is expected. 
Nevertheless, our learned modulations are designed to further augment generalizable, untrained reconstruction algorithms based on iterative optimizations. 
To validate this strategy, we deploy our learned modulations to a recently developed unsupervised iterative approach for imaging through scattering, NeuWS~\cite{feng2023neuws}, in our benchmark study.

Figure \ref{fig: NeuWS_static} shows the reconstruction performance of NeuWS~\cite{feng2023neuws} on 8 static objects, each using only 16 measurements. 
The superiority of using learned modulations is visually evident in the results. We excluded experiments with no modulations here, due to the algorithm's inability to recover objects using a single unmodulated measurement.

For dynamic scenes, each scene involves 48 frames captured by cycling through the 16 SLM patterns.
Figure \ref{fig: NeuWS_dynamic} demonstrates the reconstruction results on dynamic scenes: the result on the left illustrates a camera aperture gradually enlarging; the result on the right shows two different motions to solve the non-regional dynamic reconstruction (the number 6 on the left rotates clockwise at $0.5^{\circ}$ per frame, while the number 5 translates from the center to the right). 
Quantitative metrics on the dynamic results are included in Table \ref{tab:exp_psnr} and \ref{tab:exp_ssim}. Full video results are in the \href{https://wavemo-2024.github.io/}{project webpage}.

\begin{table}
\captionsetup{skip=5pt}
    \vspace{-5pt}
    \centering
    \begin{tabular*}{\columnwidth}{@{\extracolsep{\fill}} lcccc }

    \toprule
         \multirow{2}{*}{Method}  & \multirow{2}{*}{Data type} & \multicolumn{3}{c}{Modulation} \\
         \cline{3-5} \noalign{\vspace{1pt}}& & None & Random & Learned\\
    \midrule
         Proxy & Tissue &{16.53}  & {17.57} & \textbf{19.06} \\
         Proxy & Out-of-dist. & {9.34} & {9.90} & \textbf{10.71}\\
    \midrule
         Iterative~\cite{feng2023neuws} & Static  & {N/A} & {11.26} & \textbf{14.61}\\
         Iterative~\cite{feng2023neuws} & Dynamic & {N/A} & {8.90} & \textbf{12.89}\\
    \bottomrule
    \end{tabular*}
    \caption{\textbf{PSNR of Experimental Results.}  For our jointly trained feed-forward proxy reconstruction network (``proxy''), we tested on 40 tissue samples and 8 out-of-distribution scenes, all of which are static. For the iterative method~\cite{feng2023neuws}, we tested on the same 8 out-of-distribution static scenes. We also tested~\cite{feng2023neuws} on 2 dynamic scenes, each with 48 frames. The iterative method relies on multiple wavefront modulations and therefore cannot recover objects with a single measurement, hence the ``N/A". Compared against random modulations or no modulations, our learned modulations lead to better reconstruction performance for both the proxy network and the unsupervised iterative approach. }
    \vspace{-5pt}
    \label{tab:exp_psnr}
\end{table}

\begin{table}
\captionsetup{skip=5pt}
    \vspace{-5pt}
    \centering
    \begin{tabular*}{\columnwidth}{@{\extracolsep{\fill}} lcccc }
    \toprule
         \multirow{2}{*}{Method}  & \multirow{2}{*}{Data type}  & \multicolumn{3}{c}{Modulations} \\
         \cline{3-5}\noalign{\vspace{1pt}} & & None & Random & Learned \\
    \midrule
         Proxy & Tissue & 0.44 & 0.48 & \textbf{0.58}\\
         Proxy & Out-of-dist. & 0.29 & 0.30 & \textbf{0.32}\\
    \midrule
         Iterative~\cite{feng2023neuws} & Static  & {N/A} & 0.21 & \textbf{0.38}\\
         Iterative~\cite{feng2023neuws} & Dynamic & {N/A} & 0.23 & \textbf{0.33}\\
    \bottomrule
    \end{tabular*}
    \caption{\textbf{SSIM of Experimental Results.} Same as Table~\ref{tab:exp_psnr} but showing SSIM. Our learned modulations performs better.}
    \vspace{-5pt}
    \label{tab:exp_ssim}
\end{table}
\section{Discussion \& Conclusion}
\label{sec:discussion}

\paragraph{Role of the Proxy Reconstruction Network.}
The proxy network not only aims to reconstruct the images; its performance also indicates how good the wavefront modulations are for the restoration task.
Therefore, joint differentiable optimization of both the proxy network and the modulations works in our favor, and the experimental results indeed indicate that this is an effective way of learning useful wavefront modulations. 
Without this proxy network, we would have to somehow connect the modulation update with the performance of an iterative optimization method, which is impractical due to time and memory constraints.

\paragraph{Regularization Perspective.}
To recover missing signal frequencies due to scattering, most existing methods rely on prior information of the scene, either hand-crafted or data-driven. This work rethinks the strategy of incorporating prior knowledge: \textit{can we impose a prior over wavefront modulations?}
The goal here is to let the resulting imaging system capture more information about the target scene through scattering.
Our results validate the effectiveness of learning a prior over the wavefront modulation domain.


\paragraph{Applicability to Other Vision Tasks.}
The end-to-end learning approach can potentially benefit other vision tasks through scattering media, using a task-specific proxy network. 
For instance, if we were to do object detection or semantic segmentation through scattering media, we can replace the proxy reconstruction U-Net with a network tailored for these two tasks, and perform end-to-end training in the same way as before. \textcolor{black}{Note that we validated our proposed method on a monochromatic imaging system, which is prevalent in medical and scientific imaging. Extending it to broadband imaging tasks, e.g.,~outdoor navigation, requires the optimization of wavelength-dependent modulation patterns, which we leave for future work.}


\paragraph{Conclusion.}
We proposed a robust end-to-end learning framework integrating the optical scattering model with a proxy image reconstruction network. 
The learned wavefront modulations can both work with the proxy reconstruction network and augment a generalizable, optimization-based algorithm. 
We conducted intensive experiments to validate our approach. 
This work shows the synergy of advanced wavefront modulation techniques with cutting-edge machine learning methods, representing a significant leap forward in computer vision and optical imaging.

\subsection*{Acknowledgements}
This work was supported in part by AFOSR Young Investigator Program award no. FA9550-22-1-0208, ONR award no. N00014-23-1-2752, seed grant from SAAB, Inc., a seed grant from the UMD Brain and Behavior Institute, \textcolor{black}{NSF Expeditions award no. IIS-1730574, ONR Scattering media award no. N00014-23-1-2714, and NSF PaThs up award no. EEC-1648451}. We thank Sachin Shah for helpful discussions and Kevin F. Kelly for providing the DMD.

{
    \small
    \bibliographystyle{ieeenat_fullname}
    \bibliography{main}
}

\clearpage
\setcounter{page}{1}
\maketitlesupplementary

%
\setcounter{section}{0}

In this document, we provide additional experimental results and ablation studies. Section~\ref{supp_same}, \ref{supp_diff} and \ref{supp_dynamic} shows additional results on static in-distribution targets, static out-of-distribution targets, and dynamic targets, respectively. Section~\ref{supp_ablation} are ablation studies regarding the number and the type of modulations. Section~\ref{supp_SD} reports the standard deviation of quantitative performance on real data.

Videos of our dynamic results are shown in our project webpage at \href{https://wavemo-2024.github.io/}{https://wavemo-2024.github.io/}.



\section{Static In-distribution Scenes}
\label{supp_same}

\begin{figure}[h]
  \centering
   \includegraphics[width=0.47\textwidth]{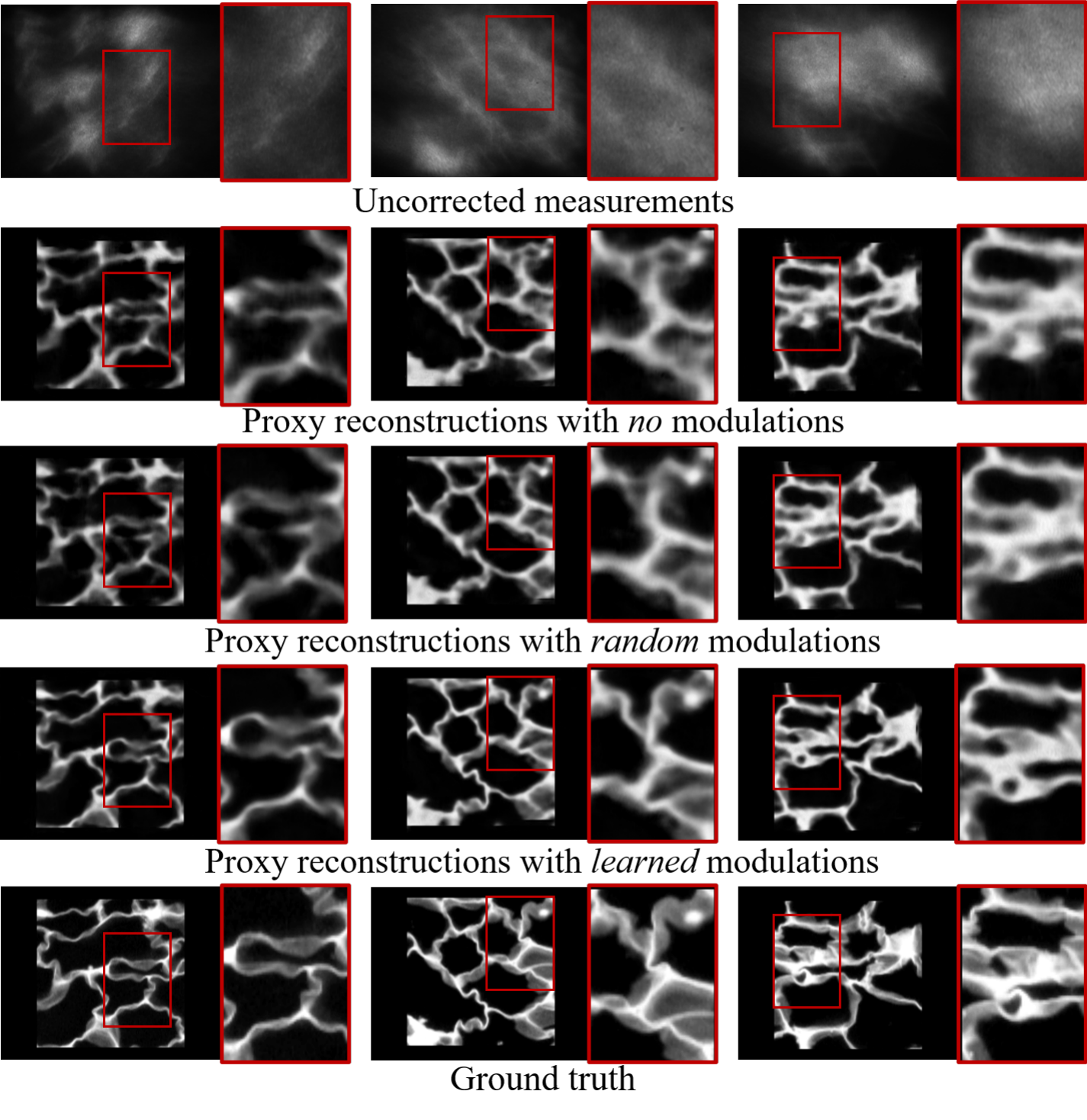} 
   \caption{\textbf{Proxy Network Reconstruction of In-distribution Targets (supplementary to Figure~\ref{fig: feedfoward} in the main paper).} Zoomed-in regions are labeled in red boxes. Reconstructions using learned modulations contain much finer details.}
   \label{fig: same_tissue}
\end{figure}

Supplementary to Figure~\ref{fig: feedfoward} in the main paper, this section provides additional evaluation results on target scenes that belong to the same type of human body tissue (adipose) as those used for training. Here, we use the proxy network for reconstruction. As can be seen from the red zoom-in boxes in Figure~\ref{fig: same_tissue}, reconstructions with learned modulations show significantly better performance. 


\section{Static Out-of-Distribution Scenes}
\label{supp_diff}
This section provides additional evaluation results on human pathological tissue slides that do not appear in the training set, including sections of parotid, stomach, and finger. Section~\ref{supp_diff_proxy} shows results using the proxy reconstruction network, while Section~\ref{supp_diff_neuws}
shows results using an unsupervised iterative approach~\cite{feng2023neuws}.

\subsection{Proxy Reconstruction Network}
\label{supp_diff_proxy}
\begin{figure*}[t]
  \centering
   \includegraphics[width=0.99\textwidth]{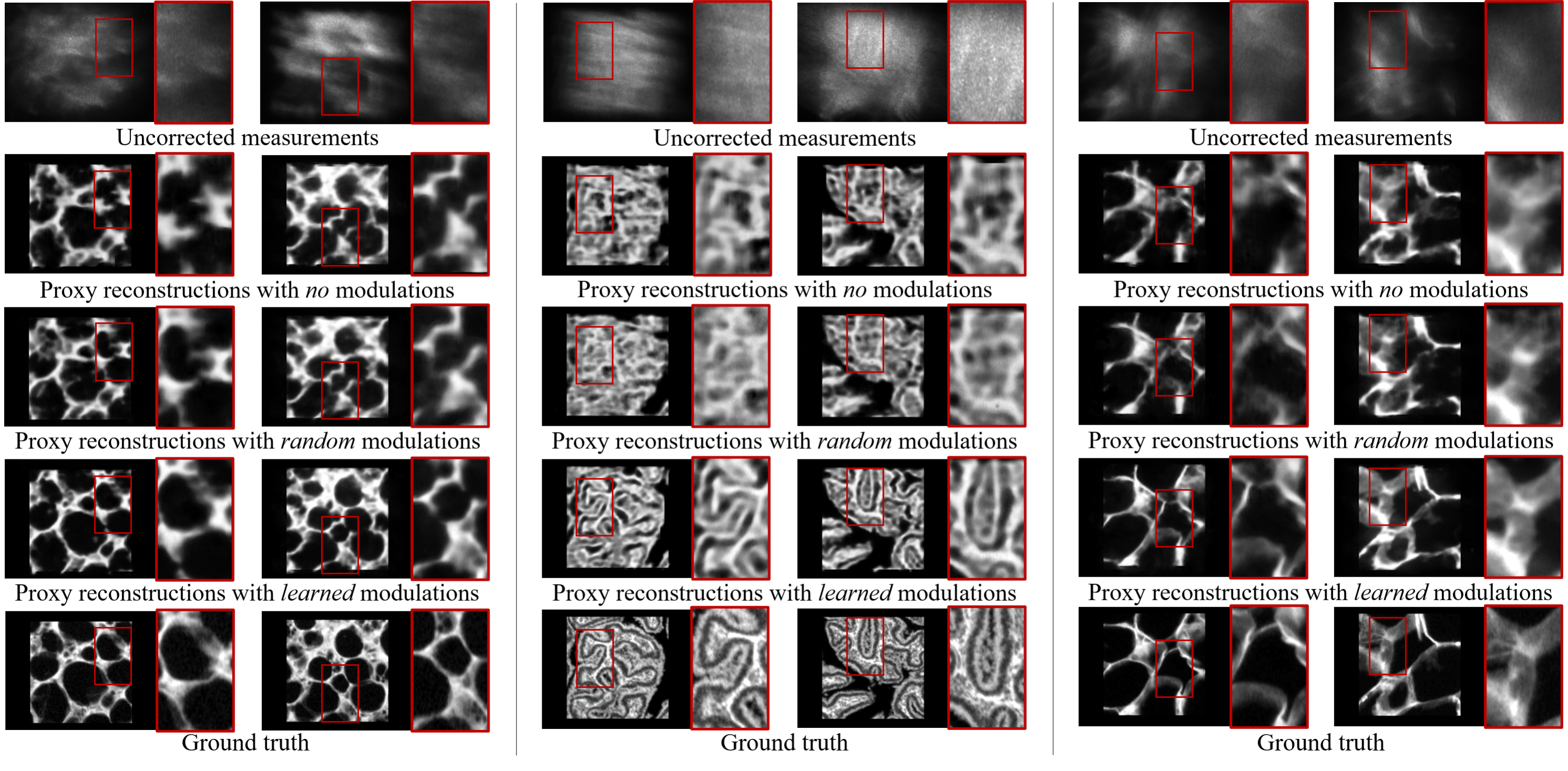} 
   \caption{\textbf{Results on Out-of-distribution Scenes using a Proxy Network Equipped with Learned Modulations.} From left to right, the 3 columns show human pathological tissue sections of parotid, stomach, and finger, respectively. Zoomed-in regions are labeled with red boxes. Learned modulations significantly enhance the reconstruction quality.}
   \label{fig: diff_tisse}
\end{figure*}

Results using a proxy reconstruction network are demonstrated in Figure~\ref{fig: diff_tisse} and Table \ref{tab:exp_diff_metric}. Reconstructions with learned modulates outperform those with random modulations and those without modulations.

\begin{table}[ht]
    \centering
    \begin{tabular}{lcccc}
    \toprule
         \multirow{2}{*}{Metric} & \multicolumn{3}{c}{Modulations} \\
         \cline{2-4}\noalign{\vspace{1pt}} & None & Random & Learned \\
    \midrule
         PSNR (SD) & 16.69 (0.37)  & 17.17 (0.36) & \textbf{18.05 (0.32)} \\
         SSIM (SD) & 0.47 (0.026) & 0.49 (0.024) & \textbf{0.56 (0.019)} \\
    \bottomrule
    \end{tabular}
    \caption{\textbf{Results on Out-of-distribution Scenes Using a Proxy Network Equipped with Learned Modulations.} The metrics are averaged over 100 samples. \textcolor{black}{We also report the standard deviation (SD) for both PSNR and SSIM.} Our learned modulations achieve the best performance.}
    \label{tab:exp_diff_metric}
\end{table}

\subsection{Unsupervised Iterative Approach}
\label{supp_diff_neuws}
\begin{figure*}[t]
  \centering
   \includegraphics[width=0.99\textwidth]{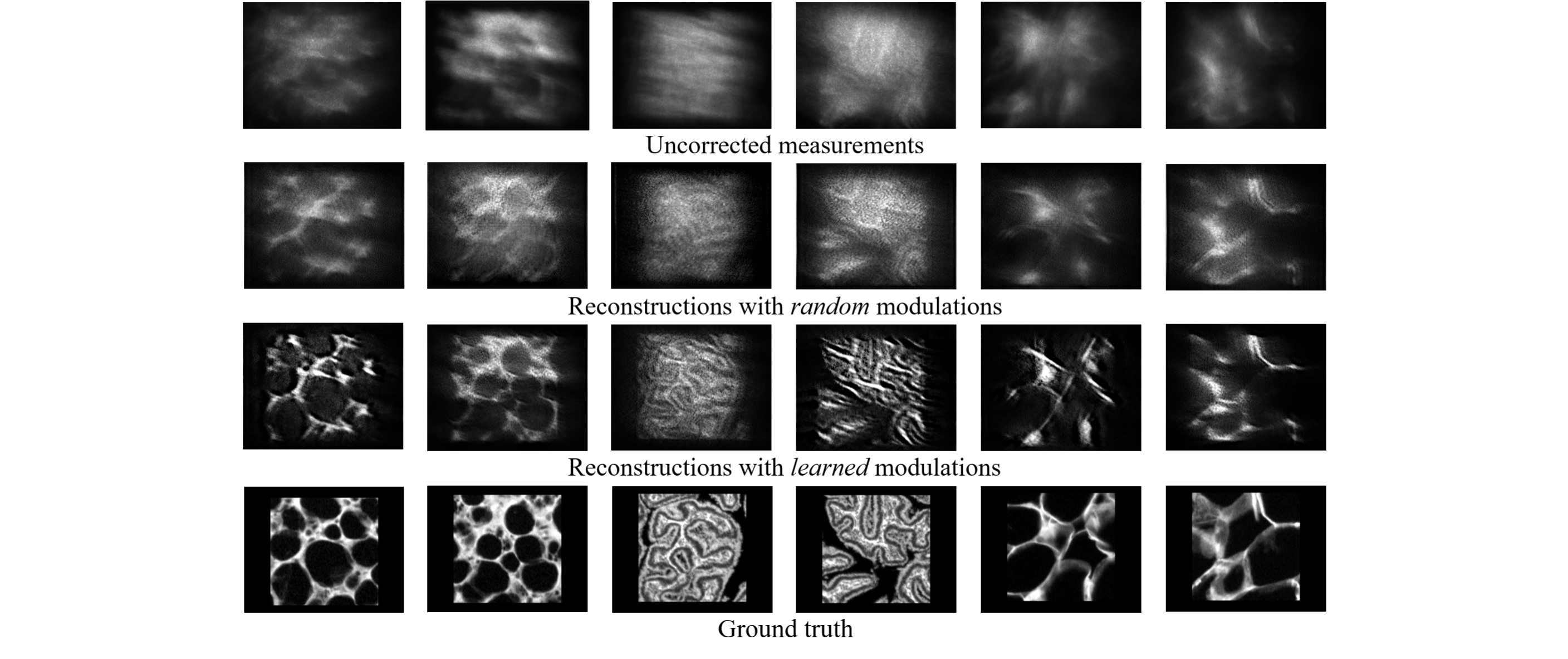} 
   \caption{\textbf{Results on Out-of-distribution Scenes Using an Unsupervised Iterative Approach~\cite{feng2023neuws} Equipped with Learned Modulations.} The target scenes include the parotid (left two columns, stomach (middle two columns), and finger (right two columns).  Reconstructions with learned modulations achieve the best quality.}
   \label{fig: neuws_diff_tisse}
\end{figure*}

Results using an unsupervised iterative approach ~\cite{feng2023neuws} are demonstrated in Figure~\ref{fig: neuws_diff_tisse} and Table~\ref{tab:neuws_diff_metric}. Similar to our previous observations, reconstructions with the 16 learned modulations exhibit clearer shapes and enhanced contrast than those with random or no modulations. 

\begin{table}[h]
    \centering
    \begin{tabular}{lcccc}
    \toprule
         \multirow{2}{*}{Metric} & \multicolumn{3}{c}{Modulations} \\
         \cline{2-4}\noalign{\vspace{1pt}} & None & Random & Learned \\
    \midrule
         PSNR (SD) & {N/A}  & 10.64 (0.78) & \textbf{12.73 (0.42)} \\
         SSIM (SD) & {N/A}  & 0.23 (0.035) & \textbf{0.32 (0.028)} \\
    \bottomrule
    \end{tabular}
    \caption{\textbf{Results on Out-of-distribution Scenes Using an Unsupervised Iterative Approach Equipped with Learned Modulations.} The metrics are averaged over six samples. \textcolor{black}{We also report the standard deviation (SD) for both PSNR and SSIM.}  The unsupervised iterative method~\cite{feng2023neuws} relies on multiple wavefront modulations and therefore cannot recover objects with a single measurement (no modulation), hence the “N/A”. Compared with random modulations or no modulations, learned modulations lead to better reconstructions.}
    \label{tab:neuws_diff_metric}
\end{table}

\newpage\thispagestyle{empty}

\begin{figure*}[htbp!]
    \centering
    \includegraphics[width=0.99\textwidth]{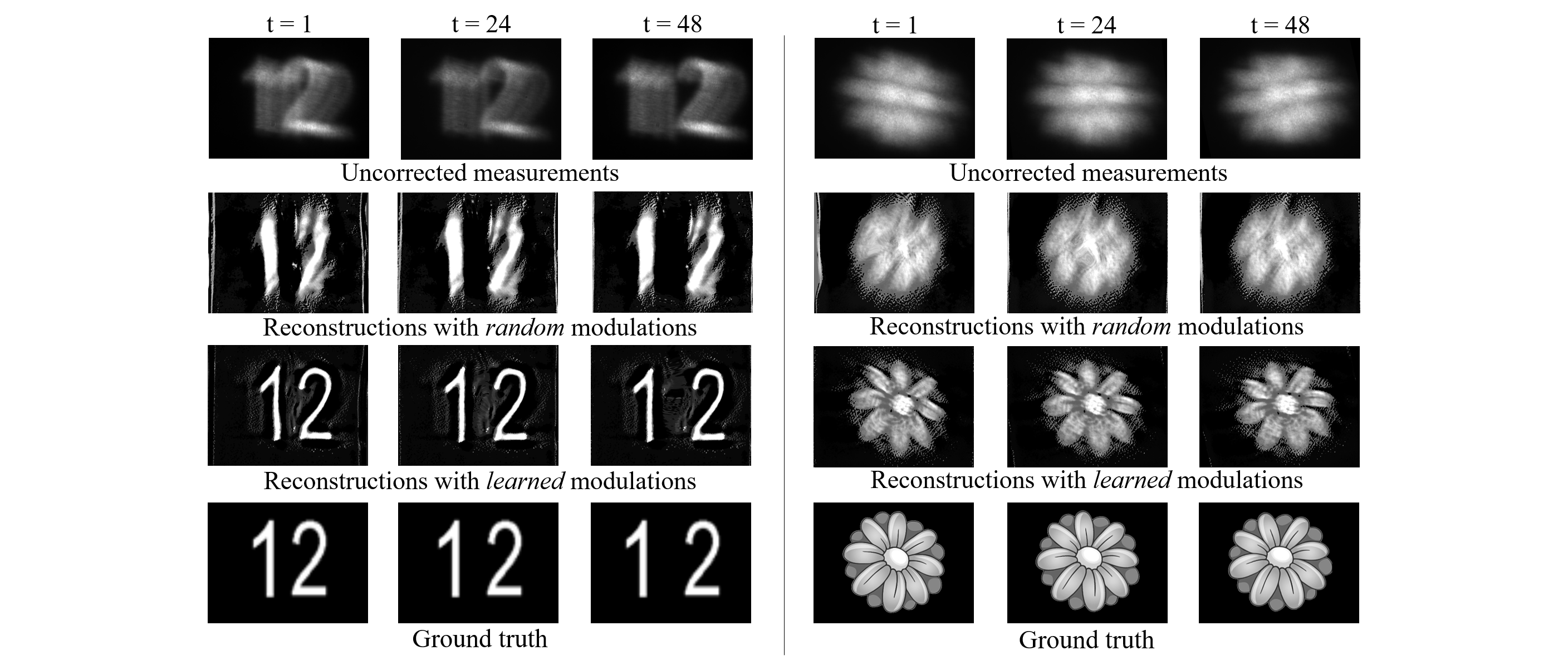} 
    \caption{\textbf{Unsupervised Reconstruction of Two Additional Dynamic Scenes. (Videos are shown in the bottom of the webpage provided in the supplementary Zip file).} In the left scene, two digits are moving in opposite directions; in the right scene, a flower is rotating. Reconstructions are done using an unsupervised iterative approach~\cite{feng2023neuws}. Our learned modulations achieve the best reconstructions.}
    \label{fig: dynamic_neuws}
\end{figure*}

\section{Dynamic Out-of-Distribution Scenes}
\label{supp_dynamic}
Supplementary to Figure~\ref{fig: NeuWS_dynamic} in the main paper, we provide two additional sets of dynamic experiments using the unsupervised iterative approach~\cite{feng2023neuws}. Same as the experiments in Figure~\ref{fig: NeuWS_dynamic}, each dynamic scene in Figure~\ref{fig: dynamic_neuws} contains 48 frames, which are captured by cycling our learned 16 modulations. In the first scene, the two digits 1 and 2 are translated in opposite directions. In the second scene, a flower undergoes a counterclockwise rotation of $0.5^{\circ}$ per frame. Even though the results obtained using learned modulations still suffer from artifacts, they are still much sharper than results using random or no modulations.
\\
\section{Ablation Studies}
\label{supp_ablation}
\subsection{Number of Modulations}
We analyze the impact of the number of modulations in simulation. As shown in Table~\ref{tab:k modulation}, PSNRs of both random (sampled from Zernike space) and learned modulations exhibit a similar trend of improvement as $K$ increases.  Note that 4 learned modulations beat 32 random modulations by over 2 dB.

\begin{table}[htbp!]
    \vspace{-5pt}
    \centering
    \fontsize{10pt}{\baselineskip}\selectfont
    \begin{tabular}{lcccc} 
    \toprule
         Modulation & $K=4$ & $K=8$ & $K=16$ &$K=32$ \\
    \midrule
         Random  & 25.550& 25.962& 26.439&  26.650\\
         Learned  & 28.996& 29.916& 30.391& 30.550\\
    \bottomrule
    \end{tabular}
    \caption{\textbf{Quality (PSNR) v.s. Number of Modulations ($K$).} PSNRs of both random (sampled from Zernike space) and learned modulations exhibit a similar trend of improvement as $K$ increases.  Note that 4 learned modulations beat 32 random modulations by over 2 dB.}
    \label{tab:k modulation}
    \vspace{-5pt}
\end{table}
\subsection{Different Types of Modulations}

\textcolor{black}{We compare our learned approach against several heuristic approaches for the design of modulation patterns, such as per-pixel random Gaussian matrices, random Zernike polynomials, focus sweeping, and directly optimizing the MTF. This comparison is done in simulation. As shown in Table~\ref{tab:heuristic_modulation}, our learned modulations notably outperform these heuristic designs. 
\begin{table}[htbp!]
\captionsetup{skip=5pt}
    \centering
    \fontsize{10pt}{\baselineskip}\selectfont
    \begin{tabular}{lccccc}
    \toprule
         PSNR & Gauss & Zern & Focus & MTF& Ours \\
    \midrule
         Mean & 26.322 & 26.439&26.096  & 25.917  & \textbf{30.391} \\
         SD & 0.38& 0.16 &  0.26 & 0.12 & 0.22\\
    \bottomrule
    \end{tabular}
    \caption{\textcolor{black}{\textbf{Quality (PSNR) v.s. Modulation Types.} We compare our learned approach against per-pixel random Gaussian matrices (Gauss), random Zernike polynomials (Zern), focus sweeping (Focus), and directly optimizing the MTF. Our approach outperforms the baselines by over 3 dB.}}
    \label{tab:heuristic_modulation}
    \vspace{-10pt}
\end{table}
}

\textcolor{black}{
\section{Standard Deviation of Results on Real Data}
\label{supp_SD}
To supplement the quantitative evaluation on real data in Tables~\ref{tab:exp_psnr} and~\ref{tab:exp_ssim}, we report the Standard Deviation (SD) as a further statistical measure, shown in Tables~\ref{tab:exp_psnr_sd} and~\ref{tab:exp_ssim_sd}.
}

\begin{table*}[htbp!]
    \vspace{-5pt}
    \centering
    \begin{tabular}{lccccc}
    \toprule
         \multirow{2}{*}{Method}  & \multirow{2}{*}{Data type} & \multicolumn{3}{c}{Modulations} \\
         \cline{3-5} \noalign{\vspace{2pt}}& & None & Random & Learned \\
    \midrule
        {Proxy} &  {Tissue} &{16.53 (0.34)}  & {17.57 (0.32)} & \textbf{19.06 (0.29)} \\
        {Proxy} & {Out-of-dist.} & {9.34 (0.38)} & {9.90 (0.37)} & \textbf{10.71 (0.35)}\\
    \midrule
        {Iterative~\cite{feng2023neuws}} & {Static}  & {N/A} & {11.26 (0.78)} & \textbf{14.61 (0.45)}\\
        {Iterative~\cite{feng2023neuws}} & {Dynamic} & {N/A} & {8.90 (0.51)} & \textbf{12.89 (0.33)}\\
    \bottomrule
    \end{tabular}
    \caption{\textbf{PSNR and Standard Deviation of Experimental Results.}  For our jointly trained feed-forward proxy reconstruction network (``proxy''), we tested on 40 tissue samples and 8 out-of-distribution scenes, all of which are static. For the iterative method~\cite{feng2023neuws}, we tested on the same 8 out-of-distribution static scenes. We also tested~\cite{feng2023neuws} on 2 dynamic scenes, each with 48 frames. The iterative method relies on multiple wavefront modulations and therefore cannot recover objects with a single measurement, hence the ``N/A". \underline{Standard deviations (SD) are included in the parentheses.} Compared against random modulations or no modulations, our learned modulations lead to better reconstruction performance for both the proxy network and an unsupervised iterative approach. }
    \vspace{-5pt}
    \label{tab:exp_psnr_sd}
\end{table*}

\begin{table*}[htbp!]
    \vspace{-5pt}
    \centering
    \begin{tabular}{lccccc}
    \toprule
         \multirow{2}{*}{Method}  & \multirow{2}{*}{Data type} & \multicolumn{3}{c}{Modulations} \\
         \cline{3-5} \noalign{\vspace{2pt}}& & None & Random & Learned \\
    \midrule
        {Proxy} & {Tissue} &{0.44 (0.023)}  & {0.48 (0.021)} & \textbf{0.58 (0.016)} \\
        {Proxy} & {Out-of-dist.} & {0.29 (0.029)} & {0.30 (0.028)} & \textbf{0.32 (0.028)}\\
    \midrule
        {Iterative~\cite{feng2023neuws}} & {Static}  & {N/A} & {0.21 (0.034)} & \textbf{0.38 (0.031)}\\
        {Iterative~\cite{feng2023neuws}} & {Dynamic} & {N/A} & {0.23 (0.032)} & \textbf{0.33 (0.029)}\\
    \bottomrule
    \end{tabular}
    \caption{\textbf{SSIM (SD) of Experimental Results.} \textcolor{black}{Same as Table~\ref{tab:exp_psnr_sd} but with SSIM. \underline{Standard deviations (SD) are included in the parentheses.} Our learned modulations lead to better performance.}}
    \vspace{-5pt}
    \label{tab:exp_ssim_sd}
\end{table*}


\end{document}